\documentclass[english]{lni}

\IfFileExists{latin1.sty}{\usepackage{latin1}}{\usepackage{isolatin1}}

\usepackage{graphicx}
\usepackage{fancyhdr}
\usepackage{array}
\usepackage{xcolor,colortbl}
\usepackage{url}
\usepackage{multirow}
\usepackage{textcomp}
\graphicspath{{./graphics/}}
\usepackage{listings} 
\usepackage{changepage} 
\usepackage[figurename=Fig., tablename=Tab., small]{caption}[2008/04/01]
\renewcommand{\lstlistingname}{List.}    

\fancypagestyle{titlepage}{
\fancyhead[RO]{\small A. Br\"omme, C. Busch, A. Dantcheva, C. Rathgeb and A. Uhl (Eds.): BIOSIG 2018, \linebreak Lecture Notes in Informatics (LNI), Gesellschaft f\"ur Informatik, Bonn 2018 \hspace{5pt} \thepage \hspace{0.05cm}} 
\fancyfoot{}}

\renewcommand{\headrulewidth}{0.4pt} 
\author{Ewelina Bartuzi \footnote{Biometrics and Machine Intelligence Laboratory, Research and Academic Computer Network, Kolska 12, 01045 Warsaw, Poland, corresponding author: ewelina.bartuzi@nask.pl} , Katarzyna Roszczewska$^1$, Mateusz Trokielewicz$^1$ and Rados\l{}aw Bia\l{}obrzeski$^1$}

\title{MobiBits: Multimodal Mobile Biometric Database}
\begin{document}

\maketitle

\renewcommand{\refname}{References}
\setcounter{footnote}{2} 
\thispagestyle{titlepage}
\pagestyle{fancy}
\fancyhead{} 
\fancyhead[RO]{\small MobiBits: Multimodal Mobile Biometric Database \hspace{5pt} \thepage \hspace{0.05cm}}
\fancyhead[LE]{\hspace{0.05cm}\small \thepage \hspace{5pt} E. Bartuzi,  K. Roszczewska, M. Trokielewicz, and R. Bia\l{}obrzeski}
\fancyfoot{} 
\renewcommand{\headrulewidth}{0.4pt} 

\begin{abstract}
This paper presents a novel database comprising representations of five different biometric characteristics, collected in a mobile, unconstrained or semi-constrained setting with three different mobile devices, including characteristics previously unavailable in existing datasets, namely hand images, thermal hand images, and thermal face images, all acquired with a mobile, off-the-shelf device. In addition to this collection of data we perform an extensive set of experiments providing insight on benchmark recognition performance that can be achieved with these data, carried out with existing commercial and academic biometric solutions. This is the first known to us mobile biometric database introducing samples of biometric traits such as thermal hand images and thermal face images. We hope that this contribution will make a valuable addition to the already existing databases and enable new experiments and studies in the field of mobile authentication. The {\it MobiBits} database is made publicly available to the research community at no cost for non-commercial purposes.
\end{abstract}

\section{Introduction}
\label{sec:Intro}
%

\vskip-3mm
Implementation of biometric authentication on mobile devices in the recent years has lead to an almost universally ubiquitous presence of biometrics in our daily lives. Starting with the fingerprint sensor of the iPhone 5s in 2013, more than 700 mobile devices have since supported this kind of authentication, with many manufacturers employing other biometric characteristics: iris in Samsung Galaxy S8 or Microsoft Lumia 950, and recently also face in 3D with Apple iPhone X's Face ID. Today, biometric authentication is used for unlocking a phone, but also for protecting financial assets in payment systems, \emph{e.g.,} Apple Pay. With this reliance upon mobile identity, secure and convenient authentication methods are crucial, hence the recently increasing interest in biometric authentication for mobile. Biometric traits such as face and fingerprint have already been given considerable attention in the scientific community, therefore, the main contributions of our paper include:\\
	{\bf (1) a multimodal biometric database -- \textit{MobiBits}, comprising representations of five different biometric characteristics, acquired with sensors embedded in commercially available, consumer-grade mobile phones,} \\
	{\bf (2) a novel use of thermal imaging enabled smartphone, adding a new dimension to typical, visible spectrum images of face and hand, offering additional cues that can be utilized, \emph{e.g.,} for Presentation Attack Detection,}\\
	{\bf (3) experiments employing commercial and academic methods, which provide benchmark results of recognition performance for the introduced datasets.}\\
According to our knowledge, our dataset offers the largest number of biometric characteristics represented in a single database, and employs the richest collection of data acquisition scenarios, which are trying to mimic those of potential real-world applications.

\vspace{-3mm}
\section{Related work}
\label{sec:Related} 
\vskip-3mm
\textbf{Iris. } 
Iris recognition, usually carried out with specialized devices operating in near infrared (NIR), poses several challenges when implemented on mobile -- most importantly, the need to utilize visible light images. Raja {\it et al.} investigated visible spectrum iris recognition on mobile devices by employing deep sparse filtering \cite{KiranSparse2014} and K-means clustering \cite{KiranKmeans}, being able to achieve equal error rate (EER) of 0.31\%. Trokielewicz {\it et al.} showed that high quality images acquired with an iPhone 5s can be successfully used with existing commercial and academic iris matchers, with FTE of 0\%, and close-to-perfect correct recognition rates, and introduced a dataset of high quality iris images \cite{TrokielewiczVisibleISBA2016}. Cross-spectral iris recognition between mobile phone samples and typical NIR images is studied in \cite{TrokielewiczBartuziJTIT}, with EER$\simeq2\%$. \\
\textbf{Handwritten signatures. }
With advantages such as ease of use, familiarity, and social acceptability, signature biometrics has the potential of mobile use. On a smartphone, one can acquire on-line handwritten signatures (X,Y) coordinates in time, as well as pressure (selected phones). On-line signature recognition methods can be divided into two groups: nonparametric and parametric. The most popular method from the first group is Dynamic Time Warping (DTW) \cite{DTWsignatureBlanco2, DTW_Putz2006}, which normalizes signatures and matches the two samples. EERs achieved in experiments with drawing tablets ranged from 0.6 \% to 3.4\% for random forgeries and from 5.4\% to 17.18\% for skilled forgeries. The most common parametric method are Hidden Markov Models \cite{HMMsignatureVan, HMMsignatureTolosana}. These methods determine the probability of sample belonging to the learned model of transitions between observation states. EERs range from 4.6\% to 7.3\% for skilled forgeries, whereas the average for random forgeries EER$\simeq$3\%.\\
\textbf{Face. }
In 2012, the bi-modal MOBIO database was published \cite{MOBIO}, including experiments for face and speaker biometrics running on a Nokia N93i smartphone. Fusing both characteristics resulted in EER$\simeq$10\%. Half error rates were achieved using the same database and a range of different methods in the ICB 2013 Face Recognition Evaluation \cite{ICB2013Face}. Mobile face recognition with locally executed implementations has recently gained interest due to the iPhone X, which employs 3D face recognition.\\
\textbf{Voice. }
Due to its convenience in some applications and increasing accuracy, text-independent speaker recognition can be a natural choice for mobile authentication systems. The bi-modal MOBIO database \cite{MOBIO}, contains a collection of voice samples captured with a Nokia N93i. This database was also used in the ICB 2013 Speaker Recognition Evaluation \cite{ICB2013Voice}, resulting in encouraging error rates of 5\% EER.\\
\textbf{Hand. }
Hand images can supply information on both hand geometry and texture. The authors of \cite{Geometry_mobile} developed a biometric recognition method based on features extracted from hand silhouette and its contour, obtaining EER=3.7\%. According to the works of Zhang {\it et al.} and Sun {\it et al.} \cite{Zhang03onlinepalmprint, Sun_2005_Palmprint}, large areas of the palmar side of the hand provide enough personal information for the identity authentication, even with low-resolution images of less than 100 dpi. Hand could be a useful for biometric authentication thanks to simple acquisition process using the built-in camera on a mobile phone without an additional sensor. The most popular feature extraction methods utilize a bank of Gabor filters \cite{Kim_palmprint_mobile, Franzgrote_palmprint_mobile}, and SIFT descriptors \cite{Choras_palmprint_mobile}, with best EER=0.79\% obtained by Kim {\it et al.} \cite{Kim_palmprint_mobile}.

\textbf{Available mobile biometric datasets. }
The {\bf MobBIO} database by Sequeira {\it et al.} \cite{MobBio_db} contains data acquired from 105 volunteers with an Asus Transformer Pad TF 300 tablet. It includes 16 voice samples, 8 eye images and 16 face images captured in different lighting conditions. The {\bf BioSecure} database by Ortega-Garcia {\it et al.} has been collected at several universities participating in the BioSecure project \cite{BioSecure_db}. It contains data collected in two acquisition sessions from 713 volunteers, each of them providing 4 frontal face images, 12 fingerprint samples, 18 voice recordings, and 25 handwritten signatures (genuine and skilled forgeries). The {\bf MBMA} database \cite{MMB_Arnowitx_db} contains three characteristics (voice, face, signature) collected in two sessions with four 4 different mobile devices. The first session was used for system development, with samples acquired from 100 volunteers. Each volunteer provided 5 face images, 3 voice records and 6 handwritten signatures. Data from the second session was acquired from 32 people and it contains 4 samples of voice recordings, 3 face images and 8 signatures. The {\bf Face-Teeth-Voice} collection introduced in \cite{KimFaceTeethVoice2010} gathers 1000 biometric samples: face images, teeth images, and voice recordings in an experiment involving 50 people, hence 20 samples per person, all obtained in a single session. Finally, the {\bf I-Am} database from \cite{AbateArmandEarMobile2017} comprises a combination of physical and behavioral biometrics in the form of ear shape images and accelerometer/gyroscope data, obtained from 100 volunteers in 1 to 3 acquisition sessions. The database collects 300 ear shape samples, and 600 accelerometer/gyroscope data samples. These datasets, including the {\bf MOBIO} dataset described in the previous paragraphs, are summarized in Table \ref{Tab: database_comp}.

\begin{table}[t!]
\renewcommand{\arraystretch}{1.0}
\centering
\footnotesize
\resizebox{\linewidth}{!}{
\begin{tabular}{>{\raggedright\arraybackslash}m{0.11\linewidth}>{\raggedright\arraybackslash}m{0.12\linewidth}>{\raggedright\arraybackslash}m{0.1\linewidth}>{\raggedright\arraybackslash}m{0.12\linewidth}>{\raggedright\arraybackslash}m{0.16\linewidth}>{\raggedright\arraybackslash}m{0.09\linewidth}>{\raggedright\arraybackslash}m{0.10\linewidth}>{\raggedright\arraybackslash}m{0.08\linewidth}} 
\hline

\textit {}  & \cellcolor{blue!30} \textbf{\emph{MobiBits}} &  \textit{MobBIO} &  \textit{MBMA} &  \textit{BioSecure} & \textit{FTV} & \textit{I-Am} & \textit{MOBIO}\\

\hline\hline
{\bf Volunteers} & \cellcolor{blue!30} 53 & 105 & 100 & 713 & 50 & 100 & 150\\\hline
{\bf Acquisition} \linebreak {\bf sessions} & \cellcolor{blue!30} 3 & 1 & 2 & 2 & 1 & 1--3 & 12\\\hline
{\bf Devices used} & \cellcolor{blue!30} Huawei Mate S \linebreak Huawei P9 Lite \linebreak CAT S60& Asus \linebreak Transformer \linebreak Pad TF 300 & iPhone 4s \linebreak Galaxy S2 \linebreak iPad 2 \linebreak Motorola Xoom & Samsung Q1 \linebreak Philips SPC900NC Webcam \linebreak HP iPAQ hx2790 PDA & HP iPAQ rw6100 & Samsung \linebreak Galaxy S4 & Nokia N93i \linebreak MacBook\\\hline
{\bf Biometric} \linebreak {\bf characteristics} \linebreak {\bf included} &  \cellcolor{blue!30} {\bf signatures}  \linebreak {\bf voice}  \linebreak {\bf face}   \linebreak {\bf iris} \linebreak {\bf hand}  & {\bf voice} \linebreak {\bf face} \linebreak {\bf iris} & {\bf voice} \linebreak {\bf signatures} \linebreak {\bf face}  & {\bf voice} \linebreak {\bf signatures} \linebreak {\bf face} \linebreak {\bf fingerprints} & \textbf{face} \linebreak \textbf{teeth} \linebreak \textbf{voice} & \textbf{arm gestures} \linebreak \textbf{ear shape} & \textbf{face} \linebreak \textbf{voice} \\
\hline
\end{tabular}}
\caption{Comparison between existing mobile multimodal biometric databases the \textit{MobiBits} database. }
\label {Tab: database_comp}
\end{table}

\vspace{-3mm}
\section{Database characteristics}
\label{sec:Database} 
\vskip-3mm
The \emph{MobiBits} multimodal mobile biometric database includes data collected from 53 volunteers (20 female and 33 male), using three different smartphones. The age of subjects ranged from 14 to 71 years. The data were collected in 3 sessions organized during 3 following months. Data collection took place twice in each acquisition session, separated by approximately 15 minutes. All samples were acquired in typical office conditions with air conditioning set to $24^o$C. Other factors that might have influenced the results, such as gender, age, health condition,  environmental conditions before the measurement, time from last meal, were recorded in the metadata. For further information on how to get access to the data please contact the authors at: \url{mobibits@nask.pl}.

Three smartphones equipped with custom data acquisition software and different technical specifications and capabilities were used for data collection: Huawei Mate S, Huawei P9 Lite, and CAT s60, with the following accessories: an Adonit Dash 2 stylus for signatures to get a more precise contact with the display, hot pillows, and cooling gel compresses were used to simulate different environmental conditions.


\begin{figure}
\centering
\includegraphics[width=\linewidth]{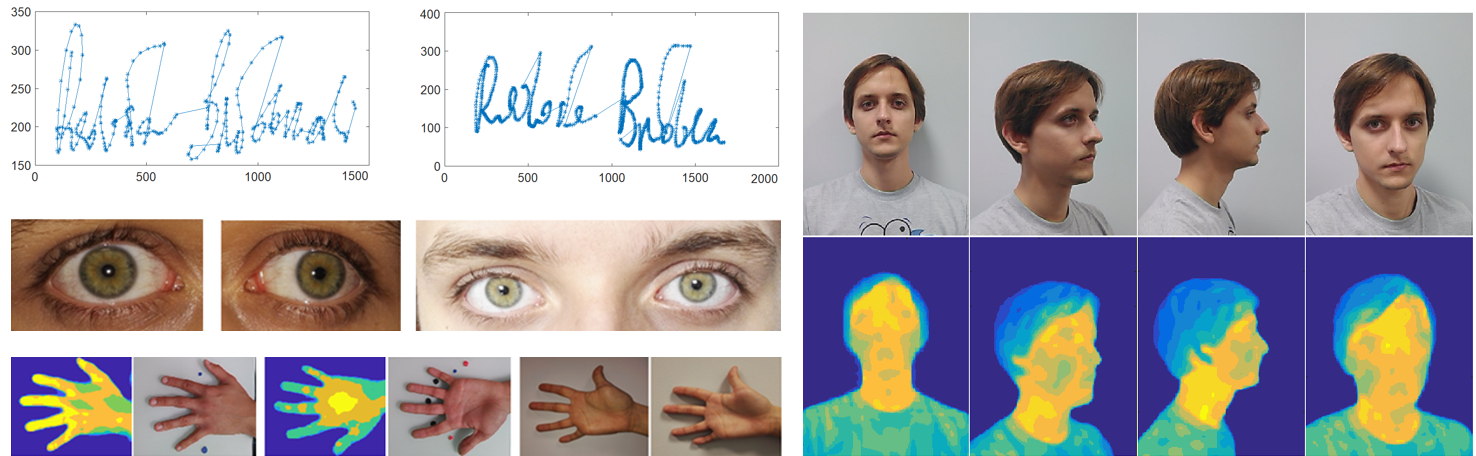}
\caption{Selected samples from the MobiBits database. {\bf Left:} genuine signatures and skilled forgeries, iris images, and hand images (thermal and visible light).  {\bf Right:} visible light and thermal face images.}.
\label{examples}
\vspace*{-0.3cm}
\end{figure}

\textbf{ON-LINE HANDWRITTEN SIGNATURES. }
The on-line handwritten signatures were recorded in real time and the order of the signature components is recorded by Huawei Mate S (with pressure-sensitive screen). Their representations contain $(x,y)$ plane coordinates and pressure values for several points on the display surface, all as a function of time. The database contains $40$ genuine signatures and $20$ skilled forgeries per person. 

\textbf{VOICE RECORDINGS. }
The voice samples were captured using a Huawei P9 Lite smartphone in a regular office environment, with no artificial noise added. All speakers were speaking Polish language. Samples were recorded during three sessions of unconstrained speech and reading (sample rate: 44.1$kHz$, duration: 0:01 - 1:10 min).

\textbf{FACE IMAGES. }
Face images were collected in three different acquisition sessions using a CAT s60 mobile phone. This smartphone allowed collection of two types of images: photos in visible light with a resolution of $480 \times 640$ pixels and thermal images of $240 \times 320$ pixels. Each session includes at least two photos for each of s six poses: frontal images, left and right semi-profile images, left and right profile images and a 'selfie' image.

\textbf{IRIS AND PERIOCULAR. }
Rear camera (13 Mpx) of the Huawei Mate S device has been used to capture images of irises (Session 1) and face focused on the eyes region (Session 2 and 3). Data collection was carried out in a typical office setting with artificial light sources. Flash was enabled during the first and the third session, and disabled during the second session to simulate different, real-world conditions of authentication.

\textbf{HAND IMAGES. }
For hand data, images of palmar and dorsal side of both left and right hand were collected. CAT s60 phone with built-in thermal camera was used for the image acquisition. According to the current state of the art, thermal images have never been used in mobile biometrics. Due to this fact, hand images were collected according to scenario which allows the temperature dynamics analyses to stimulate rich research ideas regarding this fascinating, yet largely unstudied biometric characteristic. We collected hand images: supported by glass stand and without, after cooling, after warning and with no temperature influence. One session - {\it Extra Session}, used 13 Mpx camera, for more accurate images of the hand texture. 
\vspace{-3mm}
\section{Database evaluation: tools, protocol, and performance metrics}
\label{sec:Evaluation}
\vskip-3mm
For the evaluation of the \textit{MobiBits} database we performed benchmark calculations of the recognition accuracy for all characteristics, using a proprietary PLDA/i-vector solution for voice recordings, DTW method for signatures, VeriLook for face images, IriCore for the iris images and VGG-based CNN approach for hand images. {\bf Dynamic Time Warping} (DTW) algorithm calculates an optimal warping path between two given time series is calculated, and then used for comparison \cite{DTW_Putz2006}. {\bf VeriLook} is a face recognition technology offered by Neurotechnology \cite{VeriLook}, which uses a set of robust image processing algorithms, including deep neural networks. It allows face recognition with non-frontal images, however, the default allowed rotation angles are modest. {\bf IriCore} is an iris recognition SDK developed by IriTech \cite{IriCore}, with the exact algorithm not disclosed. As this method is fine-tuned to work with iris images compliant with the ISO/IEC 19794:2011 standard \cite{ISO}, pre-processing of photos was performed with $640\times 480$ cropping and grayscale conversion of the red channel, which is said improve the visibility of the iris texture in heavily pigmented irises \cite{TrokielewiczVisibleISBA2016, TrokielewiczBartuziJTIT}. {\bf PLDA/i-vector voice recognition} is our proprietary solution based on state-of-the-art PLDA method \cite{PLDAPrince, Ivectors} with 40 mean and variance normalized MFCC features extracted from every signal frame, a 256-component, gender-independent background model, 64-dimensional total variability subspace, and 64-dimensional PLDA model. The results were normalized using cohort symmetric score normalization \cite{DehakCohorts}. Prior to feature extraction, additional spectral subtraction voice activity detection was performed to remove non-speech periods \cite{Mak2010RobustVA}. The system was trained exclusively on the MOBIO dataset. {\bf VGG-16-Hand} is a method based on the well-known convolutional network VGG-16 model \cite{VGGSimonyanCNNsForRecognition2014}, with weights fine-tuned on a dataset of hand images from the \textit{Extra Session} with flash enabled (cf. Section \ref{sec:Database}), and its bottleneck layers modified to reflect the number of output classes. 

IriCore and VeriLook engines performed quality checks prior to feature extraction and biometric template creation. Since some samples were not accepted, {\it failure-to-enroll rates (FTEs)} were calculated to express the ratio of rejected samples to all samples. Each biometric characteristic and each fusion model was tested in a verification scenario. Averaged Receiver Operating Characteristics (ROCs) and mean Equal Error Rates (EERs) were used to present the verification accuracy. These performance metrics were calculated for comparison scores obtained by performing all possible comparisons within the data, excluding the within-session ones, as these would involve comparing highly correlated data. 

Please note that thermal images of hands and those of faces are not used for comparisons here, as we are not aware of any recognition methods that would be capable of efficient processing of such samples. Instead, our intent was to provide them as datasets that are complementary to these of visible light images, so that they can be utilized for other research, such as designing robust methods for Presentation Attack Detection (PAD).  

\begin{table}[!]
\renewcommand{\arraystretch}{1.25}
\centering
\footnotesize
\resizebox{\linewidth}{!}{
\begin{tabular}{>{\raggedright\arraybackslash}m{0.09\linewidth}|>{\centering\arraybackslash}m{0.04\linewidth}>{\centering\arraybackslash}m{0.04\linewidth}>{\centering\arraybackslash}m{0.05\linewidth}|>{\centering\arraybackslash}m{0.04\linewidth}>{\centering\arraybackslash}m{0.04\linewidth}>{\centering\arraybackslash}m{0.05\linewidth}|>{\centering\arraybackslash}m{0.04\linewidth}>{\centering\arraybackslash}m{0.04\linewidth}>{\centering\arraybackslash}m{0.05\linewidth}|>{\centering\arraybackslash}m{0.04\linewidth}>{\centering\arraybackslash}m{0.04\linewidth}>{\centering\arraybackslash}m{0.05\linewidth}|>{\centering\arraybackslash}m{0.04\linewidth}>{\centering\arraybackslash}m{0.04\linewidth}>{\centering\arraybackslash}m{0.05\linewidth}} 
&    \multicolumn{3}{c|}{{\bf VOICE}} &  \multicolumn{3}{c|}{{\bf SIGNATURE}} & \multicolumn{3}{c|}{{\bf FACE}} & \multicolumn{3}{c|}{{\bf IRIS}} &  \multicolumn{3}{c}{{\bf HAND}}\\
& classes & samples & FTE & classes & samples & FTE & classes & samples & FTE & classes & samples & FTE & classes & samples & FTE\\\hline
\hline
{\bf Session 1 }& 51 & 406 & 0.00\% & 51 & 1020 & 0.00\% & 51 & 778 & 45.89\% &  49 & 497 & 14.69\%  & 49 & 2255 & 0.00\%\\
{\bf Session 2 }& 46 & 190 & 0.00\% & 42 & 420 & 0.00\% & 47 & 1005 & 42.00\% & 46 & 525 & 10.48\%    & 49 & 1508 & 0.00\%\\
{\bf Session 3 }& 48 & 624 & 0.00\% & 50 & 500 & 0.00\% & 47 & 1105 & 40.90\% & 47 & 2082 & 12.92\%    & 49 & 1740 &0.00\%\\
\hline
\end{tabular}}
\caption{A summary of the number of subjects, images and FTE rate for each characteristic.}
\label{fte}
\end{table}

\vspace{-3mm}
\section{Results and discussion}
\label{sec:Results}
\vskip-3mm
FTEs calculated for all tested characteristics are presented in Table \ref{fte}. High FTE values for the face engine may be explained with difficulties in detecting and processing profile images by the VeriLook software, as the profile and semi-profile images constituted around 60\% of the total image count. Iris enrollment errors can be a result of using mixed quality grayscale images converted from RGB images, instead of high quality NIR images recommended for the IriCore software. Other problems may be related to: iris image noise, reflections on the eye, lashes, squinted eyes. The voice recognition engine, the CNN for hand images and the DTW method did not perform any quality checks, hence zero FTEs.

\begin{figure}[h!]
\centering
\includegraphics[width=1\linewidth]{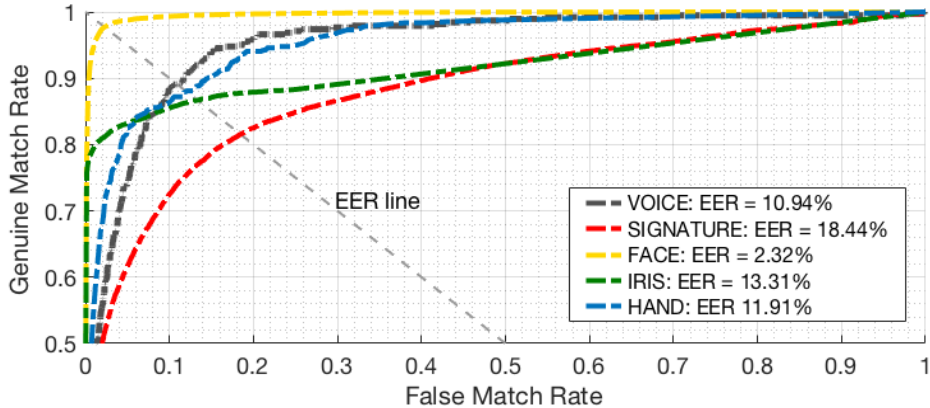}
\caption{ROC curves plotted for each individual biometric characteristic. EER values are also shown.}
\label{roc_curves}
\end{figure}

The ROC curves together with respective EER values are shown in \texttt{Figure} \ref{roc_curves}. The best single characteristic results can be achieved with face recognition, which allows verification accuracy with EER=$2.32\%$. The worst result was obtained for signatures - skilled forgeries, EER=$18.44\%$. High equal error rate for the irises is related to differences in the quality of the pictures, as high quality pictures obtained with flash enabled are compared against low-quality, no-flash pictures.

\vspace{-3mm}
\section{Conclusions}
\label{sec:Conclusions}
\vskip-2mm
The most important deliverable of this paper is the multimodal mobile biometric database -- MobiBits, comprising samples of five different biometric characteristics, collected using three different mobile devices, including a novel type of samples: thermal images of hands and faces, acquired using a thermal sensor equipped mobile phone, which we think can be a valuable contribution for studies involving presentation attack detection methods. A comprehensive set of experiments conducted on the data is reported to show the example benchmark accuracy that can be achieved on the dataset using selected commercial and academic recognition solutions. The {MobiBits} database is offered publicly at no cost for non-commercial research purposes. We hope that this will constitute a worthy addition to the mobile biometric datasets available for the biometrics community, and will provide an incentive for further research in the field of mobile biometrics. 
   
  \vspace{-5mm}
\section*{Acknowledgement}
\vskip-4mm
The authors are greatful to NASK for the  support by grant agreement no. 3/2017. The authors would like to cordially thank Dominika Szwed for her help with building the database used in this study.

\vskip-4cm

{\scriptsize
\bibliographystyle{lnig}
\bibliography{refs}
}

\end{document}